\documentclass[letterpaper]{article} 
\usepackage{aaai2026}  
\usepackage{times}  
\usepackage{helvet}  
\usepackage{courier}  
\usepackage[hyphens]{url}  
\usepackage{graphicx} 
\usepackage{natbib}  
\usepackage{caption} 
\usepackage{algorithm}
\usepackage{algorithmic}
\usepackage{amsmath}
\usepackage{amsfonts}  
\usepackage{booktabs}
\usepackage{multirow}
\usepackage{newfloat}
\usepackage{listings}
\usepackage{bibentry}
\DeclareCaptionStyle{ruled}{labelfont=normalfont,labelsep=colon,strut=off} 
\floatstyle{ruled}
\newfloat{listing}{tb}{lst}{}
\floatname{listing}{Listing}
\title{VAEVQ: Enhancing Discrete Visual Tokenization through Variational Modeling}

\author{
Sicheng Yang \equalcontrib$^{1,2}$\thanks{This work was conducted during his internship at Houmo AI.} \quad
Xing Hu \equalcontrib$^{1}$ \quad
Qiang Wu$^{1}$\quad 
Dawei Yang$^{1}$\thanks{Dawei Yang (dawei.yang@houmo.ai) is the corresponding author.} 
\\
}
\affiliations{
$^1$Houmo AI \\
$^2$Xi'an Jiaotong University\\
}

\begin{document}

\maketitle

\begin{abstract}
Vector quantization (VQ) transforms continuous image features into discrete representations, providing compressed,  tokenized inputs for generative models. However, VQ-based frameworks suffer from several issues, such as non-smooth latent spaces, weak alignment between representations before and after quantization, and poor coherence between the continuous and discrete domains. These issues lead to unstable codeword learning and underutilized codebooks, ultimately degrading the performance of both reconstruction and downstream generation tasks. To this end, we propose VAEVQ, which comprises three key components: (1) Variational Latent Quantization (VLQ), replacing the AE with a VAE for quantization to leverage its structured and smooth latent space, thereby facilitating more effective codeword activation; (2) Representation Coherence Strategy (RCS), adaptively modulating the alignment strength between pre- and post-quantization features to enhance consistency and prevent overfitting to noise; and (3) Distribution Consistency Regularization (DCR), aligning the entire codebook distribution with the continuous latent distribution to improve utilization. Extensive experiments on two benchmark datasets demonstrate that VAEVQ outperforms state-of-the-art methods. 
\footnote{Code: \url{https://github.com/script-Yang/VAEVQ}}
\end{abstract}
\section{Introduction}
Discrete visual tokenization transforms continuous image features into discrete representations by mapping them to entries in a learned codebook, typically implemented via vector quantization (VQ)~\citep{van2017neural}. In autoregressive transformers, the discrete tokens produced by VQ serve as sequential inputs for image generation~\citep{esser2021taming}, while in latent diffusion models, VQ functions as an autoencoder (AE) that defines the sampling space~\citep{rombach2022high}. Thus, the structure and utilization of the codebook are crucial to both the reconstruction quality and the expressiveness of generative models~\citep{cao2023comprehensive, tian2024visual}.

\begin{figure}[t]
\centering
\includegraphics[width=0.99\linewidth]{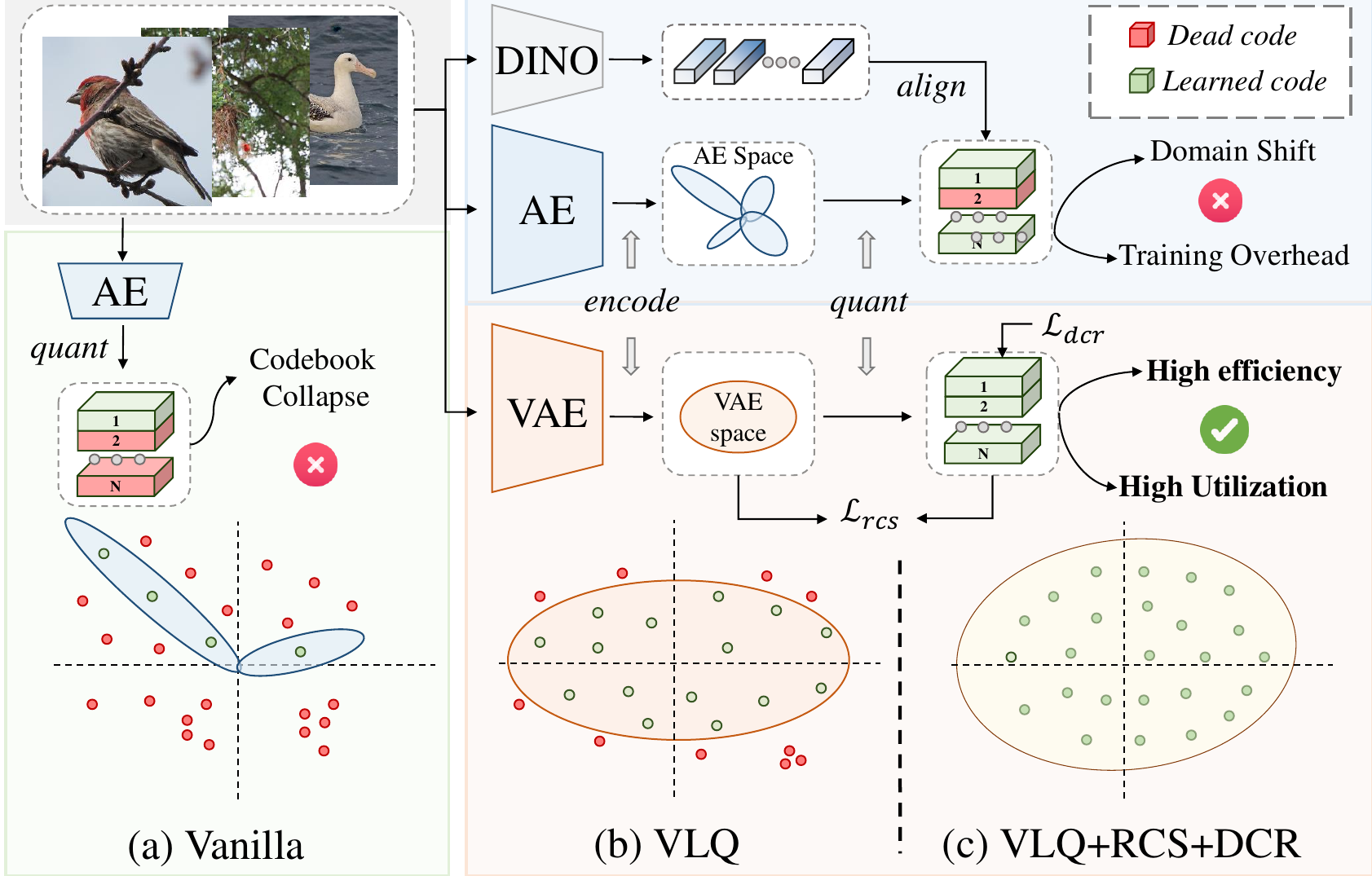}
\caption{
Comparison of different VQ strategies.  
(a) Direct quantization over AE latents often leads to codebook collapse, as the latent space of AE is typically irregular and fragmented, making it suboptimal for quantization.
(b) VLQ introduces variational modeling to smooth the transition between pre- and post-quantization representations, enabling more effective codeword activation and updating.
(c) The complete VAEVQ framework, augmented with RCS and DCR, achieves high efficiency (\textit{i.e.}, without pretrained models such as DINO) and high codebook utilization.
}
\label{fig:intro_frame}
\end{figure}

However, existing discrete visual tokenizers suffer from three major limitations. First, the latent space produced by autoencoders (AEs) is typically irregular and fragmented, forming sparse and disconnected clusters~\citep{dai2019diagnosing, vuong2023vector}. As shown in Fig.\ref{fig:intro_frame}(a), such unstructured representations hinder the effective activation and updating of codewords, eventually leading to codebook collapse. Recent methods such as FSQ~\citep{mentzer2023finite} and LFQ~\citep{yu2023language} attempt to reshape the AE latent space by forcibly compressing it and discarding its unstructured components. While this compression improves quantizability to some extent, it also introduces a representational bottleneck that significantly limits expressiveness, especially under large codebook settings~\citep{zhu2024addressing}. 

Second, the weak constraint between pre- and post-quantization representations often leads to semantic misalignment, allowing noise or unstable features to be written into the codebook. Existing methods typically minimize the distance between encoder outputs and their nearest codewords, without accounting for the noise and uncertainty in the encoder, particularly during the early stages of training~\citep{peng2021generating}. This can result in unstable codeword assignments and noisy codebook updates. VQGAN-EMA~\citep{razavi2019generating} introduces exponential moving average updates to stabilize the learning dynamics, while RQVAE~\citep{lee2022autoregressive} leverages residual connections to refine the encoded features. Nonetheless, these techniques provide only marginal improvements as the codebook size grows and cannot fundamentally resolve instability. 

Third, there is a lack of explicit structural alignment between the continuous latent space and the discrete codebook space~\cite{fang2025enhancing}. Since only a small subset of codewords is updated in each iteration, the codebook distribution may gradually drift away from the latent manifold~\citep{takida2022sq}, leaving most entries underutilized~\citep{zheng2023online}. Some methods attempt to mitigate this drift by introducing external semantic guidance. For instance, VQGAN-LC~\citep{zhu2024scaling} incorporates CLIP~\citep{radford2021learning} features, and SoftVQ~\citep{chen2025softvq} employs DINO~\citep{caron2021emerging} supervision to align token semantics. However, as illustrated in the top-right part of Fig.\ref{fig:intro_frame}, these pretrained models are trained on natural images and often suffer from domain shift when applied to fields like medical imaging~\citep{caron2021emerging}. Moreover, reliance on such external supervision introduces additional computational overhead.

In this paper, we propose VAEVQ, a unified framework composed of three key components to improve codebook utilization and representation quality in vector quantization. Specifically, We introduce the Variational Latent Quantization (VLQ) module, which performs quantization within the smooth latent space produced by a VAE, enabling more effective codeword activation and updating. We propose the Representation Coherence Strategy (RCS) to further improve representation consistency by leveraging both the encoder’s output variance and codeword information, and adaptively penalizing discrepancies between pre- and post-quantization features. We present the Distribution Consistency Regularization (DCR) module, which aligns the codebook distribution with the VAE’s Gaussian prior via optimal transport. This alignment encourages both the continuous and discrete latent spaces to conform to a shared prior, thereby mitigating global distribution mismatches. Our contributions can be summarized as follows:
\begin{itemize}

\item We propose Variational Latent Quantization (VLQ), which replaces the standard AE with a VAE for quantization. By leveraging the structured latent space and Gaussian sampling induced by the VAE, VLQ produces smoother and more organized latent features, facilitating more effective codeword activation and updating, and ultimately alleviating codebook collapse.

\item We introduce the Representation Coherence Strategy (RCS) to mitigate the semantic inconsistency between pre- and post-quantization features. RCS adaptively adjusts the alignment strength based on encoder uncertainty and codeword statistics, suppressing the influence of noisy or unstable features during codebook updates.

\item We present Distribution Consistency Regularization (DCR) to reduce global distribution mismatch between the continuous and discrete latent spaces. DCR leverages optimal transport to align the learned codebook distribution with the VAE’s Gaussian prior, promoting consistency across latent spaces and enhancing codebook utilization.

\item We conduct extensive experiments on two benchmark datasets, demonstrating that VAEVQ consistently outperforms state-of-the-art baselines in both reconstruction and generation tasks.
\end{itemize}

\section{Related Work}

\begin{figure*}[t]
\centering
\includegraphics[width=0.99\linewidth]{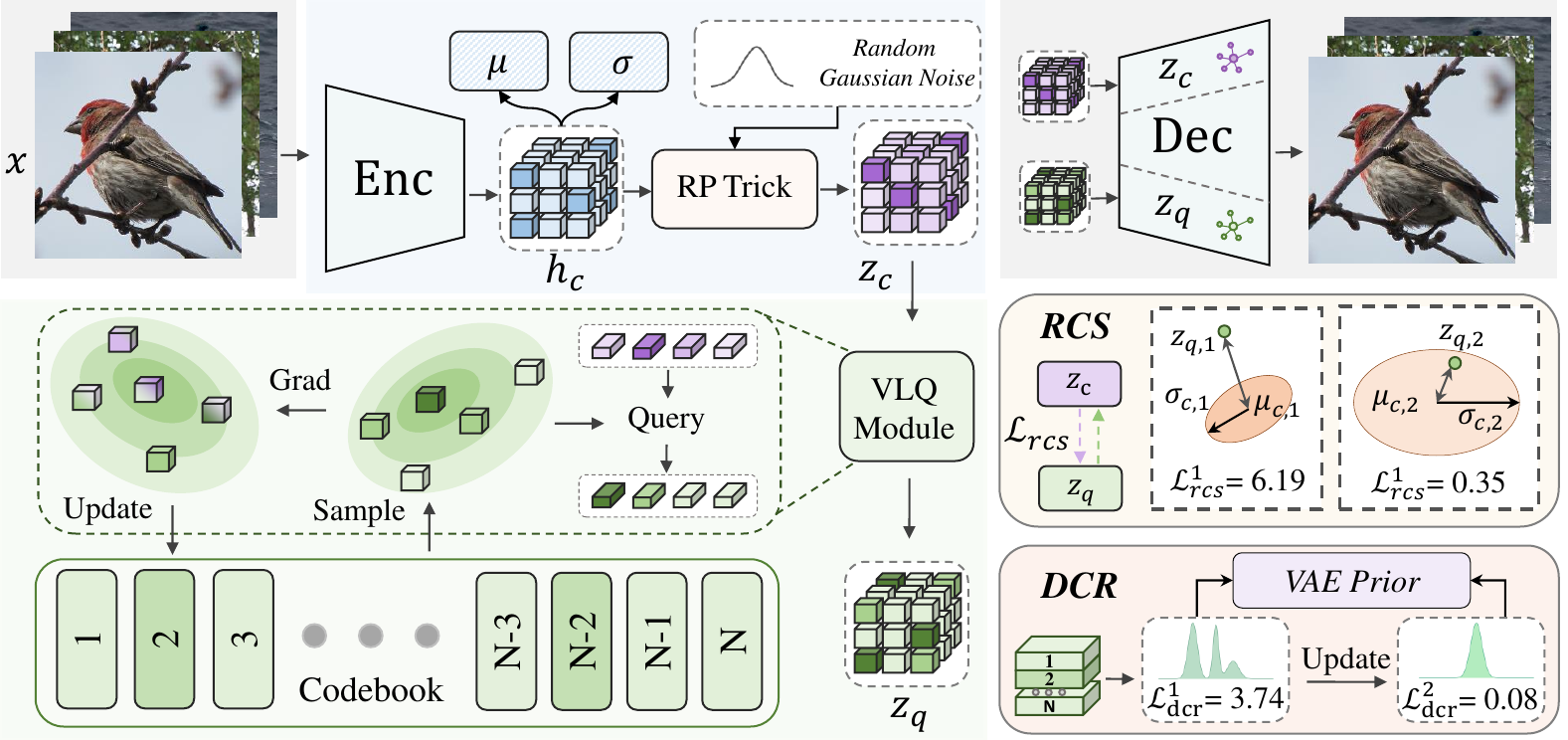}
\caption{Overview of the proposed VAEVQ framework. The VLQ module encodes the input into a variational latent vector $z_c$ and quantizes it into $z_q$, followed by dual-path decoding to enforce consistency. RCS imposes a variance-aware alignment between $z_c$ and $z_q$ to preserve confident features while tolerating uncertainty. DCR aligns the codebook distribution with the VAE prior via optimal transport. Through the joint effect of these modules, the codebook is progressively updated during training, leading to improved utilization and higher-quality visual tokens.}
\label{framework}
\end{figure*}

\paragraph{Discrete Visual Tokenizers}
Visual tokenizers convert images into compact representations for generative modeling and fall into two main categories: continuous and discrete. Continuous approaches like VAEs~\citep{kingma2013auto} offer strong semantic modeling but produce continuous outputs incompatible with token-based transformers. Discrete methods such as VQ-VAE~\citep{van2017neural} and VQGAN~\citep{esser2021taming}, as well as their numerous variants~\citep{weber2024maskbit,zhou2025onevae}, generate indexable tokens via codebooks, enabling autoregressive and diffusion models, yet often suffer from codebook collapse and semantic loss~\citep{ma2025unitok,yangvq,han2025infinity,zhang2025quantize}. To address this, we propose VAEVQ, a framework that combines the semantic richness of VAEs with the discrete structure required for token-based generation.

\subsection{Visual Tokenizers for Image Generation}
The representations produced by discrete visual tokenizers serve as the foundation for downstream tasks. In autoregressive settings~\citep{chang2022maskgit,sun2024autoregressive}, transformers predict the next token in a sequence, while latent diffusion models~\citep{rombach2022high,karras2022elucidating,esser2024scaling} iteratively denoise tokens in a learned latent space. In both cases, tokens produced by the trained codebook critically affects generation quality~\citep{razavi2019generating}. However, conventional discrete tokenizers often yield poorly utilized or semantically inconsistent codebooks. To address this, we propose VAEVQ, which combines variational encoding with vector quantization to produce discrete tokens that are both expressive and well-distributed. This unified design enhances token quality and yields improvements across diverse generative paradigms.

\section{Methodology}
\subsection{Overview}
Figure ~\ref{framework} illustrates the proposed VAEVQ framework. The VLQ module quantizes the latent space of a VAE and employs dual-path decoding to reconstruct the input from both the sampled and quantized representations. RCS adaptively aligns the pre- and post-quantization vectors at the feature level, guided by the encoder’s output variance and the corresponding codewords. DCR regularizes the global codebook distribution to match the VAE prior via optimal transport, thereby encouraging comprehensive codeword activation. Together, these components enhance codebook utilization and token quality.

\subsection{Variational Latent Quantization (VLQ)}
Traditional vector quantization (VQ) frameworks typically operate on the latent space of deterministic autoencoders (AEs). 
Although AEs can preserve local semantics to some extent through reconstruction training, their latent representations often exhibit irregular global geometry and non-uniform density~\citep{dai2019diagnosing}. 
That is, the relative distribution of latent vectors does not faithfully reflect the relative similarity structure of their corresponding inputs, which leads to distortions in the latent space and ultimately hampers quantization effectiveness~\citep{peng2021generating}. 
Such misalignment between the latent manifold and the input data manifold limits codebook utilization and may cause codeword collapse under large-scale settings. In contrast, the latent space induced by a variational autoencoder (VAE) is explicitly regularized to follow a smooth prior distribution, resulting in more continuous, semantically coherent representations that are better suited for quantization.

\begin{figure}[t]
\centering
\includegraphics[width=0.99\linewidth]{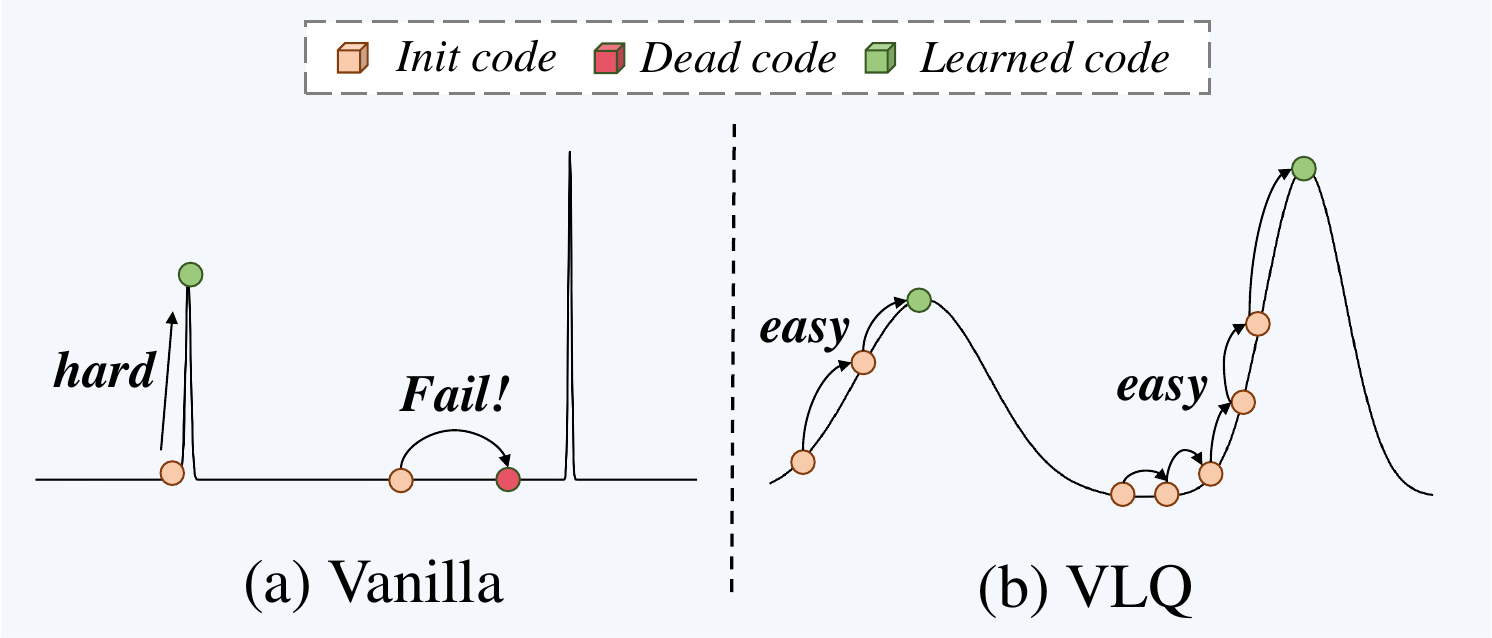}
\caption{
Comparison between vanilla vector quantization and our proposed Variational Latent Quantization (VLQ).
(a) In vanilla VQ, latent features from the autoencoder (AE) latent space are sparse and rigid, causing most initial codewords (orange) to remain unused. As a result, many codewords become inactive (red), and only a few (green) are eventually trained, leading to low codebook utilization.
(b) In VLQ, latent vectors are drawn from the VAE latent space, which has a smoother distribution. This enables more codewords to be activated and gradually updated.
}
\label{fig:vlq_framework}
\end{figure}

To overcome these limitations, we propose \textit{Variational Latent Quantization (VLQ)}. as illustrated in Fig.~\ref{fig:intro_frame}, Fig.~\ref{fig:vlq_framework}, VLQ performs quantization over latent vectors sampled from the VAE latent space, whose smoother and more continuous structure facilitates codeword learning and leads to higher codebook utilization.

Given an input image $x$, the encoder $E(\cdot)$ produces a hidden feature $h_c = E(x)$, which generates the mean $\mu_c$ and log-variance $\log \sigma_c^2$ of a diagonal Gaussian posterior $q(z|x)$. A latent vector is sampled using the reparameterization trick~\citep{kingma2013auto}:
\begin{align}
z_c = \mu_c + \sigma_c \odot \epsilon, \quad \epsilon \sim \mathcal{N}(0, I),
\end{align}
and quantized via nearest-neighbor lookup in a learnable codebook:
\begin{align}
k^* = \arg\min_k \|z_c - e_k\|_2^2, \quad 
z_q = e_{k^*}.
\end{align}
Both $z_c$ and $z_q$ are passed through a shared decoder $D$ to reconstruct the input:
\begin{align}
\hat{x}_c = D(z_c), \quad \hat{x}_q = D(z_q),
\end{align}
and the reconstruction loss is defined as:
\begin{align}
\mathcal{L}_{\text{rec}} = \|x - \hat{x}_c\|_2^2 + \|x - \hat{x}_q\|_2^2.
\end{align}

VLQ offers three key advantages. First, it quantizes latent features sampled from the VAE latent space, which tends to be more continuous and well-structured than that of AE latent space, leading to more effective and robust quantization. Second, the stochasticity introduced by $\epsilon \sim \mathcal{N}(0, I)$ encourages the sampled latent vector $z_c$ to explore the latent space more broadly, promoting diverse codeword activation and alleviating early-stage codebook collapse. Third, VLQ employs a dual-path reconstruction strategy where both $z_c$ and its quantized version $z_q$ are used for decoding. This setup not only reduces semantic drift by allowing $z_c$ to provide corrective feedback to $z_q$, but also enables $z_q$ to guide $z_c$, gradually aligning its distribution with the codebook topology and making it more quantization-friendly.

\subsection{Representation Coherence Strategy (RCS)}
Vector quantization (VQ) models often incorporate feature-level alignment objectives to bridge the gap between the continuous latent $z_c$ and its quantized counterpart $z_q$. A common approach is to apply an $\ell_2$ penalty, \textit{i.e.}, $\|z_q - z_c\|^2$, which we refer to as hard alignment. 

However, since the encoder inevitably introduces noise during training, the $\ell_2$ loss penalizes all deviations equally, regardless of whether the discrepancy is semantically meaningful or caused by uncertainty. 
This indiscriminate treatment can cause the model to overcorrect dimensions that are inherently high-variance and naturally fluctuating. 

To address these issues, we propose the \textit{Representation Coherence Strategy (RCS)}, a soft alignment mechanism that adapts the alignment strength according to the encoder's confidence in each latent dimension. In our VLQ framework, $z_c$ is not a deterministic point but a sample drawn from a Gaussian distribution parameterized by the encoder: $z_c \sim \mathcal{N}(\mu_c, \mathrm{diag}(\sigma_c^2))$. Within this formulation, the variance $\sigma_c^2$ reflects the encoder’s uncertainty. A lower variance indicates higher confidence, suggesting that the corresponding dimension encodes stable and reliable semantics. Accordingly, $z_q$ is expected to lie within this high-confidence region and deviations from it should be penalized more strongly. In contrast, higher variance implies uncertainty, and $z_q$ should be allowed to explore a wider range of plausible alternatives.

Formally, we express this behavior using the log-likelihood of $z_q$ under the distribution of $z_c$:
\begin{align}
\log p(z_q) = -\frac{1}{2} \sum_{i=1}^d \left[ \left( \frac{z_{q,i} - \mu_{c,i}}{\sigma_{c,i}} \right)^2 + \log(2\pi \sigma_{c,i}^2) \right],
\end{align}
and define the coherence loss as the negative log-likelihood:
\begin{align}
\mathcal{L}_{\text{rcs}} = -\log p(z_q).
\end{align}
To stabilize training and avoid gradient explosion in high-uncertainty regions, we detach the variance term $\sigma_c$ from the computational graph and omit the constant term, yielding the simplified objective:
\begin{align}
\mathcal{L}_{\text{rcs}} = \frac{1}{2} \sum_{i=1}^d \left( \frac{z_{q,i} - \mu_{c,i}}{\texttt{detach}(\sigma_{c,i})} \right)^2.
\end{align}

By minimizing $\mathcal{L}_{\text{rcs}}$, RCS enforces a confidence-aware soft alignment that adaptively constrains $z_q$ based on the encoder’s reliability. Dimensions with low variance, indicative of high confidence, are aligned more tightly to preserve critical semantics, while high variance dimensions are granted greater flexibility to avoid overfitting noise and to explore a wider range of codewords. This variance-guided constraint acts as a divide-and-conquer strategy, encouraging $z_c$ and $z_q$ to move closer in a targeted manner. As a result, RCS preserves essential semantic information while promoting more balanced and effective codebook utilization.
\subsection{Distribution Consistency Regularization (DCR)}
Traditional vector quantization (VQ) methods typically lack explicit constraints on the global structure of the codebook, often leading to codebook collapse or severe underutilization, where only a small subset of codewords are frequently updated during training. VLQ and RCS partially alleviate this issue: VLQ leverages a well-structured variational latent space for quantization, promoting diversified codeword activation, while RCS imposes instance-level alignment between the sampled latent $z_c$ and its quantized counterpart $z_q$ to enhance local consistency. However, neither method explicitly regulates the overall distribution of the codebook. As a result, they remain insufficient to ensure meaningful utilization of all codewords throughout training.

To address this, we introduce \textit{Distribution Consistency Regularization (DCR)}, which enforces global consistency between the discrete and continuous latent spaces.
In VAE frameworks, only the continuous latents are regularized toward the standard Gaussian prior $\mathcal{N}(0, I)$.
DCR extends this constraint to the quantized branch by encouraging the codebook embeddings to follow the same prior, formulated as a distribution alignment problem solved via optimal transport.

From a statistical perspective, the approximate Gaussianity of the quantized representations can be explained by the bounded and finite-variance nature of the latent space: when a large number of latent samples are aggregated, their empirical distribution tends to approach a multivariate Gaussian according to the central limit theorem~\citep{rosenblatt1956central,kwak2017central}. 
Therefore, we model the codebook $\mathcal{C} = \{ e_k \}_{k=1}^{K}$ as a finite set of samples drawn from an empirical Gaussian distribution:
\begin{align}
\hat{q}(z_q) &= \mathcal{N}(\mu_q, \Sigma_q), \\
\mu_q &= \frac{1}{K} \sum_{k=1}^{K} e_k, \\
\Sigma_q &= \frac{1}{K-1} 
\sum_{k=1}^{K} (e_k - \mu_q)(e_k - \mu_q)^\top.
\end{align}
This formulation characterizes the geometry of the codebook using its first- and second-order statistics, 
providing a compact parametric approximation of its global distribution.
\begin{figure}[b]
  \centering
  \includegraphics[width=0.99\linewidth]{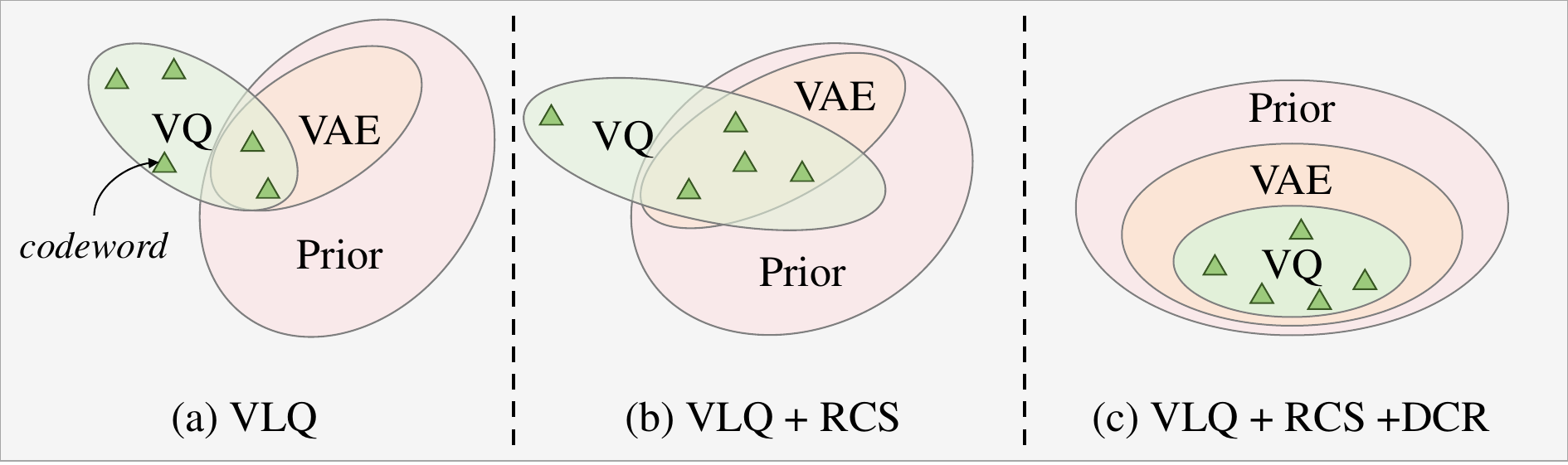}
  \caption{
Conceptual illustration of the progressive alignment among the VQ space, continuous latent space (VAE), and the prior distribution. 
(a) VQ and VAE are partially aligned, but both remain misaligned with the prior. (b) RCS encourages instance-level alignment between VQ and VAE, reducing their local discrepancies. However, some regions of the latent space remain unaligned.
(c) DCR regularizes the codebook distribution to match the Gaussian prior, yielding a diverse and well-structured codebook whose space is aligned with both the VAE latent space and the prior.
}
  \label{fig:vanilla}
\end{figure}

To align the codebook distribution $\hat{q}(z_q)$ with the VAE prior $\mathcal{N}(0, I)$, we formulate this task as an optimal transport (OT) problem. In the special case where both source and target distributions are Gaussians, the 2-Wasserstein distance~\citep{panaretos2019statistical} admits a closed-form expression. The resulting regularization objective is given by:
\begin{align}
\mathcal{L}_{\mathrm{dcr}} = \|\mu_q\|_2^2 + \mathrm{Tr}(\Sigma_q) - 2 \cdot \mathrm{Tr}(\Sigma_q^{1/2}),
\end{align}
where $\mathrm{Tr}(\cdot)$ denotes the matrix trace operator. This regularization encourages the global structure of the codebook to match the Gaussian prior, thereby improving compatibility with the variational latent space. Under the VAE framework, the continuous latent space is naturally regularized toward $\mathcal{N}(0, I)$; by minimizing $\mathcal{L}_{\mathrm{dcr}}$, the discrete codebook is similarly guided to adopt this structure.

As training progresses, codewords are dynamically adjusted to align with the distributional structure of the continuous latent space. This structural consistency facilitates smoother transitions between $z_c$ and $z_q$, reduces quantization error, and enhances codebook utilization. The improved alignment also activates a broader range of codewords, thereby increasing the expressive capacity of the codebook.

While RCS promotes instance-level consistency between $z_c$ and $z_q$, DCR complements it by globally regularizing the distribution of codebook embeddings. As illustrated in Fig.~\ref{fig:vanilla}, VLQ explicitly quantizes samples from the continuous latent space of a VAE, enabling a smoother transition to discrete representations. RCS enforces feature-level consistency between the sampled and quantized latents, reducing semantic drift caused by quantization. DCR further regularizes the overall codebook distribution by aligning it with the VAE prior, promoting balanced codeword activation. Together, these components not only stabilize the learning dynamics but also maintain a well-structured latent space that facilitates effective and balanced codeword usage, thereby enhancing codebook utilization.

\subsection{Training Objective}

To ensure a fair comparison, we adopt the same encoder and decoder architecture as VQGAN~\citep{esser2021taming}. The overall objective is formulated as:
\begin{align}
\mathcal{L}_{\text{total}} =
\mathcal{L}_{\text{rec}} +
\lambda_{\text{rcs}} \mathcal{L}_{\text{rcs}} +
\lambda_{\text{dcr}} \mathcal{L}_{\text{dcr}} +
\lambda_{\text{net}} \mathcal{L}_{\text{net}}.
\end{align}
Here, $\mathcal{L}{\text{net}}$ includes the perceptual and adversarial losses commonly used in VQGAN~\citep{esser2021taming}, along with the KL loss from the VAE branch~\citep{van2017neural}. The weights of each loss component are determined through extensive empirical studies to ensure stable training and optimal performance. Specifically, we set $\lambda_{\text{rcs}} = 1.0$ and $\lambda_{\text{dcr}} = 0.1$, while $\lambda_{\text{net}}$ remains consistent with the default setting in VQGAN.

\section{Experiments}
\subsection{Datasets and Implementation Details}
\paragraph{Datasets.} We evaluate our method on two benchmark datasets: ImageNet~\citep{deng2009imagenet} and BraTS24~\citep{de20242024}. Both datasets are resized to $256 \times 256$. ImageNet is a large-scale natural image dataset with diverse object categories, and we follow its standard train/test split. To assess the generalization ability of our model across domains and modalities, we further include BraTS24, a medical imaging dataset that differs significantly from ImageNet in both visual appearance and semantic structure. BraTS24 contains multi-contrast 3D brain MRI scans; to ensure compatibility with our 2D framework, we extract axial slices from the volumetric data. We use 80\% of the subjects for training and the remaining 20\% for evaluation and testing. Both datasets are used for reconstruction and generation tasks, enabling a comprehensive evaluation of VAEVQ's performance across diverse visual domains.
\paragraph{Implementation Details.}
We use VQGAN~\citep{esser2021taming} as the primary baseline and compare it with several representative variants and competing methods, including Mo-VQGAN~\citep{zheng2022movq}, VQGAN-EMA~\citep{razavi2019generating}, VQGAN-LC~\citep{zhu2024scaling}, SimVQ~\citep{zhu2024addressing}, SoftVQ~\citep{chen2025softvq}, and our proposed VAEVQ. 

For all models, the latent dimensionality is set to 64 and the codebook size is fixed at 16,384 for consistency and fair comparison.  All models are implemented using the PyTorch 2.4.1 framework. Training is performed from scratch for 50 epochs with a batch size of 32, using the Adam optimizer with an initial learning rate of $1 \times 10^{-4}$, following a cosine annealing schedule, on 8 NVIDIA A6000 GPUs. Performance is evaluated using three standard metrics: PSNR, SSIM, and reconstruction FID (rFID).
\subsection{Visual Reconstruction Performance}

As shown in Table~\ref{tab:cmp_tokenizers}, our proposed VAEVQ consistently outperforms all baselines and state-of-the-art methods across both datasets and all metrics. In particular, compared to the strongest prior method SimVQ, VAEVQ achieves a 0.03dB improvement in PSNR, 2\% in SSIM, and a 0.72 reduction in rFID on ImageNet. On BraTS24, it brings a further 2.09dB PSNR gain, 0.02 SSIM improvement, and a 1.86 drop in rFID. These results underscore the superior reconstruction quality and robust domain generalization capability of our approach. Notably, although VQGAN-LC attempts to leverage external pre-trained feature extractors for enhanced tokenization, it exhibits a relatively high rFID on the BraTS24 dataset (9.78), suggesting that such domain-agnostic priors may induce semantic drift when applied to medical images with significantly different visual structures.
\begin{table}[t]
\centering
\fontsize{9}{11}\selectfont 
\setlength{\tabcolsep}{4pt}  
\begin{tabular}{lccc|ccc}
\toprule
\multirow{2}{*}{Method} & \multicolumn{3}{c|}{ImageNet} & \multicolumn{3}{c}{BraTS24} \\
\cmidrule(lr){2-4} \cmidrule(lr){5-7}
 & PSNR↑ & SSIM↑ & rFID↓ & PSNR↑ & SSIM↑ & rFID↓ \\
\midrule
VQGAN       & 19.28 & 0.53 & 8.02 & 21.59 & 0.81 & 10.47 \\
VQGAN-EMA   & 20.23 & 0.55 & 6.36 & 23.37 & 0.83 & 9.35  \\
Mo-VQGAN    & 21.12 & 0.57 & 4.49 & 25.33 & 0.85 & 8.91  \\
VQGAN-LC    & 21.48 & 0.62 & 3.52 & 26.59 & 0.89 & 9.78  \\
SoftVQ      & 21.73 & 0.65 & 2.03 & 27.34 & 0.91 & 5.12  \\
SimVQ       & 22.02 & 0.66 & 1.86 & 29.82 & 0.93 & 4.36  \\
\textbf{Ours} & \textbf{22.05} & \textbf{0.68} & \textbf{1.14} & \textbf{31.91} & \textbf{0.95} & \textbf{2.50} \\
\bottomrule
\end{tabular}
\caption{Comparison of visual tokenizers on ImageNet and BraTS24 using PSNR, SSIM, and rFID. Higher PSNR/SSIM and lower rFID are better.}
\label{tab:cmp_tokenizers}
\end{table}

\begin{figure}[t]
\centering
\includegraphics[width=0.99\linewidth]{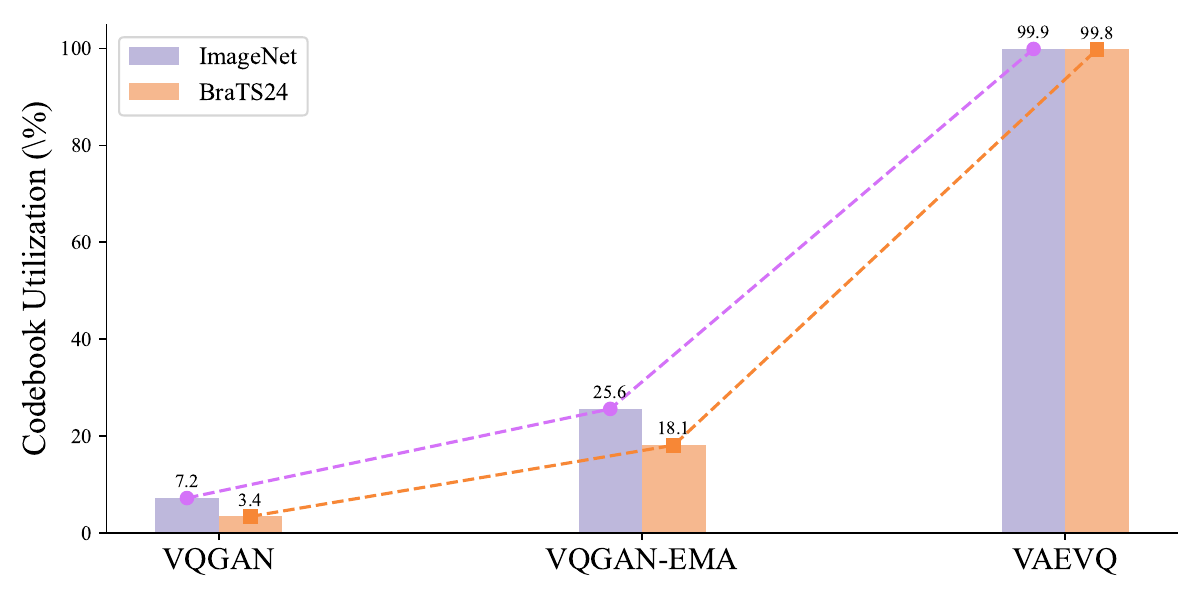}
\caption{
Codebook utilization rates (\%) of different tokenizers on ImageNet and BraTS24. VAEVQ achieves significantly higher utilization across both datasets, indicating more effective and diverse token usage.
}
\label{fig:utils}
\end{figure}

We further analyze the codebook utilization rates~\citep{tian2024visual} on the two benchmark datasets,  as illustrated in Fig.~\ref{fig:utils}. The baseline VQGAN shows significant underutilization, with only 7.2\% and 3.4\% of the codebook entries effectively used on the two datasets, indicating that most codewords remain inactive. Although its EMA-based variant achieves moderate improvements, the overall utilization remains suboptimal. In stark contrast, our VAEVQ activates nearly all codebook entries, effectively resolving the issue of insufficient codeword usage.

\begin{figure}[t]
\centering
\includegraphics[width=0.99\linewidth]{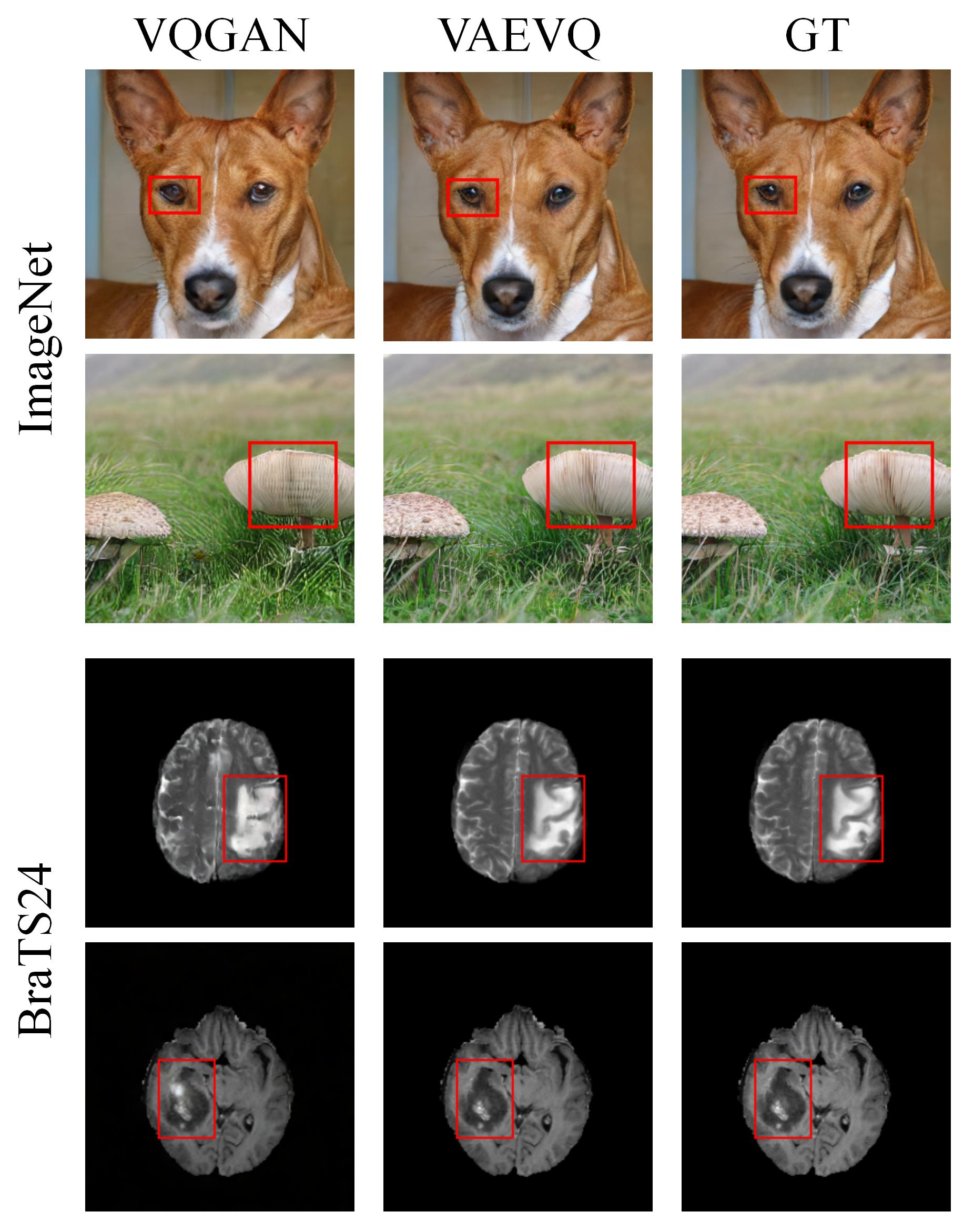}
\caption{
Visual comparison of reconstructed images on two benchmark datasets. Compared to existing methods, VAEVQ achieves superior reconstruction quality with sharper textures and enhanced structural preservation.
}
\label{fig:vis_recon}
\end{figure}

Fig.~\ref{fig:vis_recon} presents qualitative comparisons of reconstructed samples from both datasets. 
Compared to existing methods, VAEVQ generates reconstructions with clearer textures, sharper boundaries, and stronger semantic consistency. These visual results, combined with the quantitative evaluations and codebook analysis, demonstrate the effectiveness of VAEVQ in achieving high perceptual quality while faithfully preserving structural and semantic details. More qualitative results and extended comparisons can be found in the appendix.

\subsection{Visual Generation Performance}
To evaluate the generative capability of our learned codebook, we integrate it into two mainstream generation paradigms: autoregressive and diffusion-based models. For the autoregressive setting, we adopt LlamaGen-B~\citep{sun2024autoregressive} as the generative backbone. For the diffusion-based setting, we adopt the LDM-4 model~\citep{rombach2022high}. Except for replacing the visual tokenizer with our proposed VAEVQ, all other configurations follow the original implementation. We evaluate generation quality on ImageNet and BraTS24 using the standard generation FID (gen FID) metric. As shown in Table~\ref{tab:generation_results}, our VAEVQ consistently outperforms VQGAN in terms of generative quality, achieving lower gFID scores across both generation architectures and datasets. Specifically, with LlamaGen-B, VAEVQ reduces gFID from 5.43 to 4.68 on ImageNet ($\downarrow$0.75) and from 7.54 to 4.42 on BraTS24 ($\downarrow$3.12). Similarly, under the LDM-4 backbone, VAEVQ lowers gFID from 3.60 to 2.98 on ImageNet ($\downarrow$0.62) and from 6.85 to 3.11 on BraTS24 ($\downarrow$3.74). These improvements can be attributed to the effectiveness of VAEVQ’s three core components, which jointly contribute to better generative quality across diverse settings.

\subsection{Ablation Study}
\begin{table}[t]
\centering
\fontsize{9}{11}\selectfont 
\setlength{\tabcolsep}{4pt} 

\begin{tabular}{lcccc}
\toprule
\multirow{2}{*}{Tokenizer} & \multicolumn{2}{c}{LlamaGen-B} & \multicolumn{2}{c}{LDM-4} \\
\cmidrule(lr){2-3} \cmidrule(lr){4-5}
 & ImageNet & BraTS24 & ImageNet & BraTS24 \\
\midrule
VQGAN   & 5.43 & 7.54 & 3.60 & 6.85 \\
VAEVQ   & 4.68 & 4.42 & 2.98 & 3.11 \\
\midrule
$\Delta$ (↓) & 0.75 & 3.12 & 0.62 & 3.74  \\
\bottomrule
\end{tabular}
\caption{Generation FID (gFID $\downarrow$) comparison between VQGAN and our VAEVQ on ImageNet and BraTS24 using LlamaGen-B and LDM-4. Lower values indicate better generative quality.}
\label{tab:generation_results}
\end{table}

\begin{figure}[t]
\centering
\includegraphics[width=0.99\linewidth]{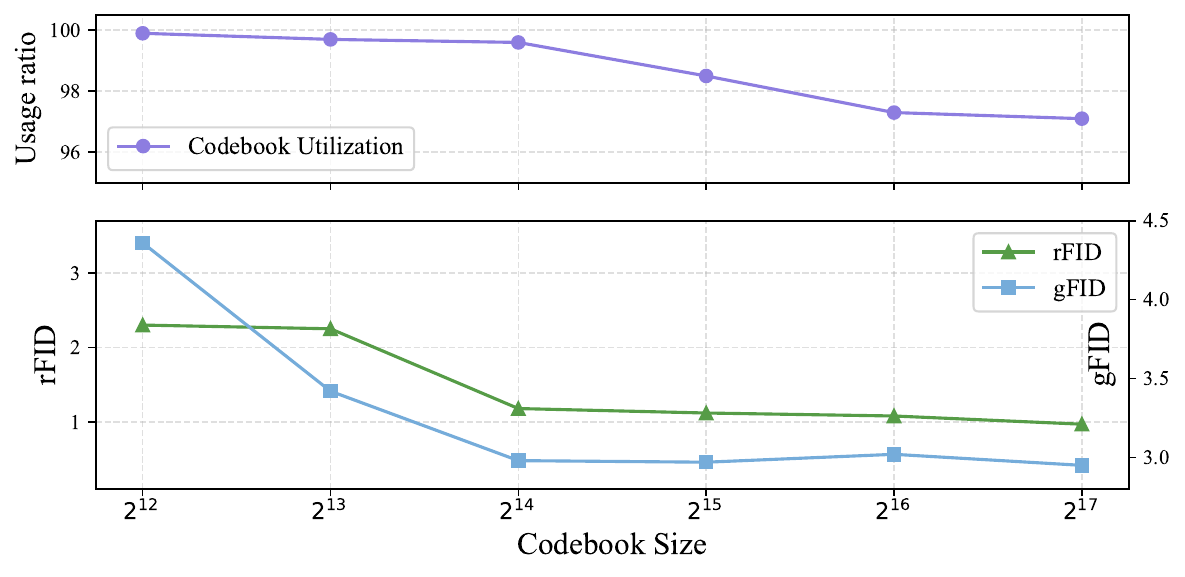}
\caption{Effect of codebook size on performance. The top plot shows the codebook utilization ratio (\%) across different codebook sizes, indicating how effectively the quantized space is used. The bottom plot reports the rFID and gFID. }
\label{fig:codebook_ablation}
\end{figure}

\paragraph{Impact of Codebook Size.} 
We further investigate the impact of codebook size on reconstruction quality, generation performance, and codebook utilization on the ImageNet dataset. As shown in Fig.~\ref{fig:codebook_ablation}, we conduct experiments using VAEVQ under varying codebook sizes ranging from 4096 ($2^{12}$) to 131,072 ($2^{17}$). The generation model is based on the LDM-4 architecture. Our results indicate that increasing the codebook size generally improves reconstruction fidelity, as a larger dictionary offers finer granularity for encoding visual details. However, we observe that the generation performance tends to saturate once the codebook size exceeds 16,384, likely due to increased token entropy and sparsity. Therefore, we use a codebook size of 16,384 to balance reconstruction quality and generation stability.

Moreover, we observe consistently high codebook utilization (above 95\%) across different sizes, indicating that our framework scales well and avoids codebook collapse even at large scales.
\begin{table}[t]
\centering
\fontsize{9}{11}\selectfont 
\setlength{\tabcolsep}{4pt}  
\begin{tabular}{l|ccc|cc}
\toprule
Method & VLQ & RCS & DCR & ImageNet & BraTS24 \\
\midrule
Baseline         &              &              &              & 8.02 & 10.47 \\
M1            & \checkmark   &              &              & 2.84 & 5.12 \\
M2            & \checkmark   & \checkmark   &              & 1.92 & 2.98 \\
M3            & \checkmark   &              & \checkmark   & 2.18 & 3.26 \\
Ours (Full)   & \checkmark   & \checkmark   & \checkmark   & \textbf{1.14} & \textbf{2.50} \\
\bottomrule
\end{tabular}
\caption{Component-wise ablation on ImageNet and BraTS24 using rFID, where lower values indicate better reconstruction quality. We evaluate the individual and combined contributions of VLQ, RCS, and DCR modules.}
\label{tab:ablation}
\end{table}

\paragraph{Impact of Modular Components.}
We conduct ablation studies to evaluate the individual and combined contributions of VLQ, RCS, and DCR, as summarized in Table~\ref{tab:ablation}. Starting from the standard VQGAN as the baseline (rFID: 8.02 on ImageNet and 10.47 on BraTS24), we progressively incorporate each proposed module.

Introducing VLQ alone (M1), which quantizes the latent space of a variational autoencoder, results in a significant reduction in rFID, with values decreasing to 2.84 on ImageNet and 5.12 on BraTS24. This demonstrates the advantage of learning a smoother and more structured latent space. Incorporating RCS (M2), which enforces instance-level coherence between the sampled latent $z_c$ and its quantized counterpart $z_q$, further improves performance, reducing the rFID to 1.92 on ImageNet and 2.98 on BraTS24. Replacing RCS with DCR (M3), which encourages alignment between the codebook distribution and the VAE prior, also yields favorable results, achieving rFID scores of 2.18 and 3.26, respectively. When all three modules are combined, the full model achieves the best performance, with the lowest rFID of 1.14 on ImageNet and 2.50 on BraTS24. These findings indicate that VLQ serves as the primary driver of performance improvement, while RCS and DCR provide complementary benefits by enhancing feature-level alignment across the quantization boundary and promoting global consistency between the codebook and the continuous latent space. Together, these components contribute to improved codebook utilization and reconstruction quality.

\section{Conclusion}

In this paper, we propose VAEVQ, a framework that enhances vector quantization for visual representation learning. VAEVQ introduces three key components: Variational Latent Quantization (VLQ), which performs quantization over a structured and smooth latent space learned through a VAE; Representation Coherence Strategy (RCS), which adaptively adjusts the alignment strength between pre- and post-quantization features to improve local consistency; and Distribution Consistency Regularization (DCR), which aligns the global distribution of codebook embeddings with the latent prior to promote codebook utilization. Extensive experiments on two benchmark datasets demonstrate that VAEVQ consistently outperforms previous methods in terms of reconstruction quality, generative fidelity, and codebook efficiency, without relying on pretrained models.

While VAEVQ demonstrates strong empirical performance, its use of a fixed-size codebook may constrain flexibility when dealing with data of varying complexity. In future work, we plan to investigate adaptive codebook scaling strategies that allow the model to dynamically adjust the codebook size during training.

\appendix

\nobibliography*


\clearpage
\onecolumn
\begin{center}
    {\LARGE \textbf{Supplementary Material for}}\\[0.8em]
    {\Large \textbf{VAEVQ: Enhancing Discrete Visual Tokenization through Variational Modeling}}\\[1em]
    {\normalsize Sicheng Yang$^{1,2}$, Xing Hu$^{1}$, Dawei Yang$^{1,}$, Qiang Wu$^{1}$}\\[0.5em]
    {\small $^{1}$Houmo AI \quad $^{2}$Xi'an Jiaotong University}\\[2em]
\end{center}

\setcounter{section}{0}
\renewcommand{\thesection}{A.\arabic{section}}

\subsection{Quantization and Vector Quantization}
Quantization is a fundamental technique for converting continuous-valued representations into discrete forms, thereby reducing the computational cost and memory footprint of neural networks. This discretization enables efficient storage and computation using low-bit integers while approximating the behavior of high-precision floating-point arithmetic. Such low-bit quantization has been widely employed in large-scale models to accelerate inference without significantly compromising accuracy~\citep{yang2024post,hu2025ostquant,hu2024llm,hu2025moequant}.

In contrast, vector quantization (VQ) extends this concept to a higher-dimensional latent space by replacing scalar quantization with the assignment of entire feature vectors to the nearest code entries. Specifically, given an encoder output $z_e(x) \in \mathbb{R}^d$, the quantized representation is obtained as
\begin{equation}
    z_q(x) = e_k, \quad \text{where } k = \arg\min_i \|z_e(x) - e_i\|_2^2,
\end{equation}
where $\{e_i\}_{i=1}^K$ represents the learnable codebook that contains $K$ discrete embedding vectors. This operation effectively partitions the latent space into Voronoi cells, each corresponding to one code entry. During training, the encoder and the codebook are jointly optimized through a commitment loss that encourages the encoded features to stay close to their assigned code vectors, ensuring stable learning dynamics. 

Unlike post-training quantization in large language models~\citep{xu2025rsavq,xu2025mambaquant,xu2025rwkvquant}, which primarily serves to compress pre-trained weights, vector quantization in VAEVQ acts as a discrete information bottleneck. It allows the model to learn symbolic, high-level representations that capture semantic regularities, bridging the gap between continuous neural features and discrete token-based modeling. This property makes it powerful building blocks for modern discrete generative frameworks and vision-language tokenizers.

\subsection{Derivation and Implementation of Distribution Consistency Regularization}

\paragraph{Closed-form Wasserstein Distance between Gaussians}
Given two multivariate Gaussians $\mathcal{N}_1 = \mathcal{N}(\boldsymbol{\mu}_1, \boldsymbol{\Sigma}_1)$ and $\mathcal{N}_2 = \mathcal{N}(\boldsymbol{\mu}_2, \boldsymbol{\Sigma}_2)$, the squared 2-Wasserstein distance between them is~\citep{panaretos2019statistical}:
\begin{align}
\mathcal{W}_2^2 \left( \mathcal{N}_1, \mathcal{N}_2 \right)
= \| \boldsymbol{\mu}_1 - \boldsymbol{\mu}_2 \|_2^2 
+ \operatorname{Tr}(\boldsymbol{\Sigma}_1 + \boldsymbol{\Sigma}_2) 
- 2 \cdot \operatorname{Tr}(\mathbf{A}),
\end{align}
where $\boldsymbol{\mu}_1, \boldsymbol{\mu}_2$ are the means of the two Gaussians,  
$\boldsymbol{\Sigma}_1, \boldsymbol{\Sigma}_2$ are their covariance matrices,  
$\operatorname{Tr}(\cdot)$ denotes the matrix trace, and  
$\mathbf{A} = \left( \boldsymbol{\Sigma}_1^{1/2} \boldsymbol{\Sigma}_2 \boldsymbol{\Sigma}_1^{1/2} \right)^{1/2}$ is the geometric mean of the covariances in the Wasserstein space.

In the special case where the second distribution is the standard Gaussian $\mathcal{N}_0 = \mathcal{N}(\mathbf{0}, \mathbf{I})$, and the source is $\mathcal{N}_1 = \mathcal{N}(\boldsymbol{\mu}, \boldsymbol{\Sigma})$, the expression simplifies to:
\begin{align}
\mathcal{W}_2^2 \left( \mathcal{N}_1, \mathcal{N}_0 \right)
= \| \boldsymbol{\mu} \|_2^2 + \operatorname{Tr}(\boldsymbol{\Sigma}) 
- 2 \cdot \operatorname{Tr}(\boldsymbol{\Sigma}^{1/2}).
\end{align}
Here, $\boldsymbol{\Sigma}^{1/2}$ denotes the symmetric matrix square root, \textit{i.e.}, the unique positive semi-definite matrix satisfying $\boldsymbol{\Sigma}^{1/2} \boldsymbol{\Sigma}^{1/2} = \boldsymbol{\Sigma}$.
\paragraph{Application to Codebook Regularization}
In our setting, we estimate the empirical distribution of the codebook embeddings as a Gaussian:
\begin{align}
\boldsymbol{\mu}_q &= \frac{1}{K} \sum_{k=1}^K \mathbf{e}_k, \\
\boldsymbol{\Sigma}_q &= \frac{1}{K - 1} \sum_{k=1}^K (\mathbf{e}_k - \boldsymbol{\mu}_q)(\mathbf{e}_k - \boldsymbol{\mu}_q)^\top,
\end{align}
where $\{ \mathbf{e}_k \}_{k=1}^K$ are the $K$ codebook vectors. The DCR loss is then defined as:
\begin{align}
\mathcal{L}_{\text{dcr}} = \| \boldsymbol{\mu}_q \|_2^2 
+ \operatorname{Tr}(\boldsymbol{\Sigma}_q) 
- 2 \cdot \operatorname{Tr}(\boldsymbol{\Sigma}_q^{1/2}),
\end{align}
where $\boldsymbol{\mu}_q$ and $\boldsymbol{\Sigma}_q$ are the empirical mean and covariance of the codebook, and $\boldsymbol{\Sigma}_q^{1/2}$ is the symmetric matrix square root.

\subsection{More Visualization}
\begin{figure*}[t]
\centering
\includegraphics[width=0.95\linewidth]{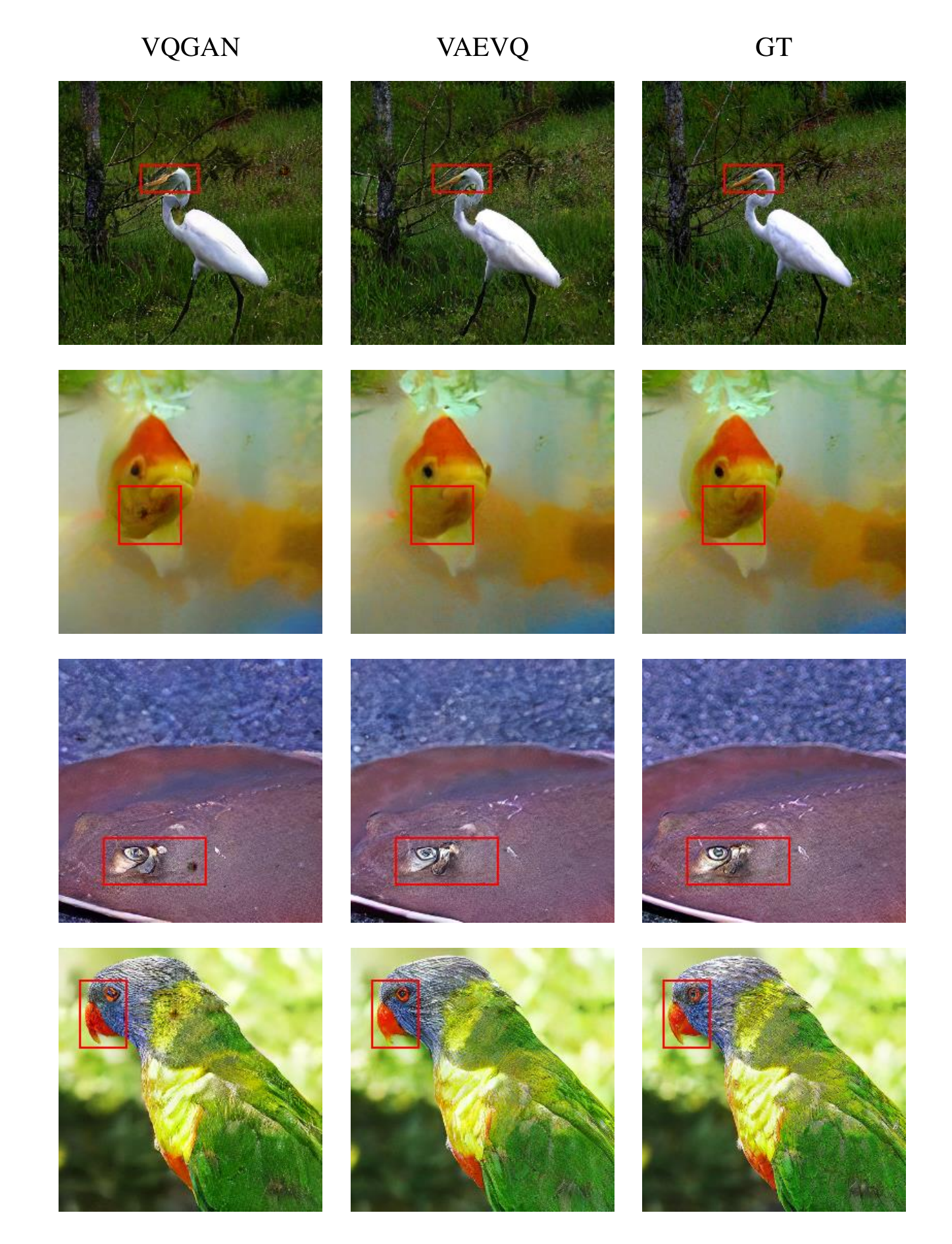}
\caption{Visual comparison of reconstructed images on the ImageNet dataset.}
\label{fig:imagenet_appendix}
\end{figure*}
\begin{figure*}[t]
\centering
\includegraphics[width=0.95\linewidth]{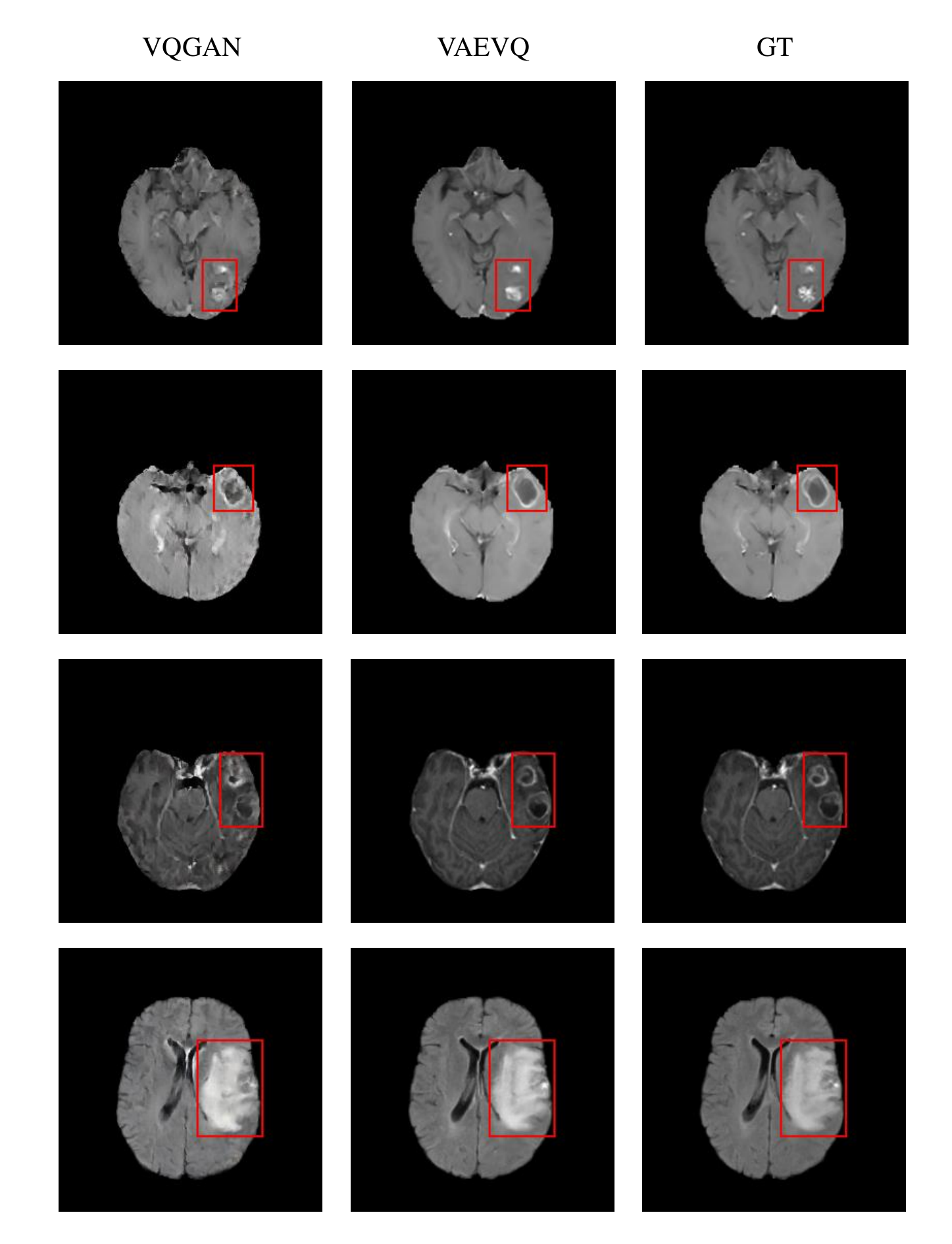}
\caption{Visual comparison of reconstructed images on the BraTS dataset.}
\label{fig:brats_appendix}
\end{figure*}

We present additional reconstruction results on both the ImageNet and BraTS datasets. As shown in Fig.~\ref{fig:imagenet_appendix} and Fig.~\ref{fig:brats_appendix}, VAEVQ consistently reconstructs images with clearer textures, sharper edges, and improved structural coherence compared to baseline methods. These qualitative results complement the main paper by visually confirming the advantages of our proposed VLQ, RCS, and DCR modules across both natural and medical imaging domains.

\subsection{Extending VAEVQ in the Future Work}
As a further elaboration on the future directions discussed in the main text, we explore the potential of extending VAEVQ with \textit{hierarchical codebooks} or \textit{multi-resolution quantization schemes}, which can further enhance the model's flexibility, expressiveness, and adaptability to diverse data characteristics.

In particular, hierarchical codebooks enable the model to represent information at multiple semantic levels. A coarse-level codebook can efficiently capture global structure and high-level semantics, while fine-level codebooks can encode local textures and detailed variations. Such multi-scale quantization facilitates compact yet expressive representations and allows the model to selectively focus on important regions during reconstruction.

One promising approach is to organize the codebook into a tree-structured hierarchy~\citep{xu2025rsavq}, where each coarse codeword is linked to a set of finer-grained sub-codewords. During inference, the model can traverse this hierarchy based on input-specific uncertainty or task-specific guidance, enabling content-aware quantization with dynamic resolution control.

Another direction involves a mixture-of-experts design, where multiple quantization modules operate at different granularities. A learned controller or gating mechanism can dynamically select or combine outputs from these modules, adapting the quantization process to spatial or semantic complexity across regions. This design improves representational efficiency while maintaining reconstruction fidelity, especially for heterogeneous inputs.

\end{document}